\documentclass[a4paper]{ifacconf}

\usepackage{dblfloatfix}
\usepackage{epsfig}
\usepackage{psfrag,graphicx,color}
\usepackage[normalem]{ulem}
\usepackage{amsmath,amsfonts,amssymb}
\usepackage{epsfig,psfrag,graphicx,color}
\usepackage{algpseudocode}
\usepackage{algorithm}
\usepackage{color} 

\font\bfmath=cmmib10
\mathchardef\Gamma="7100
\mathchardef\Delta="7101
\mathchardef\Theta="7102
\mathchardef\Lambda="7103
\mathchardef\Xi="7104
\mathchardef\Pi="7105
\mathchardef\Sigma="7106
\mathchardef\Upsilon="7107
\mathchardef\Phi="7108
\mathchardef\Psi="7109
\mathchardef\Omega="710A

\mathchardef\alpha="710B
\mathchardef\beta="710C
\mathchardef\gamma="710D
\mathchardef\delta="710E
\mathchardef\epsilon="710F
\mathchardef\zeta="7110
\mathchardef\eta="7111
\mathchardef\theta="7112
\mathchardef\iota="7113
\mathchardef\kappa="7114
\mathchardef\lambda="7115
\mathchardef\mu="7116
\mathchardef\nu="7117
\mathchardef\xi="7118
\mathchardef\pi="7119
\mathchardef\rho="711A
\mathchardef\sigma="711B
\mathchardef\tau="711C
\mathchardef\upsilon="711D
\mathchardef\phi="711E
\mathchardef\chi="711F
\mathchardef\psi="7120
\mathchardef\omega="7121
\mathchardef\epsilon="7122

\mathchardef\varepsilon="7122
\mathchardef\vartheta="7123
\mathchardef\varpi="7124
\mathchardef\varrho="7125
\mathchardef\varsigma="7126
\mathchardef\varphi="7127
\mathchardef\imath="717B
\mathchardef\jmath="717C

\def\zero{{\bf 0}}
\def\Zero{{\mbox{\boldmath $O$}}}

\def\bff{{\mbox{\boldmath $f$}}}
\def\bfg{{\mbox{\boldmath $g$}}}
\def\bfh{{\mbox{\boldmath $h$}}}

\def\bfp{{\mbox{\boldmath $p$}}}
\def\bfq{{\mbox{\boldmath $q$}}}

\def\bfx{{\mbox{\boldmath $x$}}}
\def\bfy{{\mbox{\boldmath $y$}}}

\def\bfC{{\mbox{\boldmath $C$}}}
\def\bfD{{\mbox{\boldmath $D$}}}

\def\bfF{{\mbox{\boldmath $F$}}}

\def\bfI{{\mbox{\boldmath $I$}}}
\def\bfJ{{\mbox{\boldmath $J$}}}
\def\bfK{{\mbox{\boldmath $K$}}}

\def\bfM{{\mbox{\boldmath $M$}}}

\def\bfbeta{{\mbox{\boldmath $\beta$}}}
\def\bfgamma{{\mbox{\boldmath $\gamma$}}}

\def\bfsigma{{\mbox{\boldmath $\sigma$}}}
\def\bftau{{\mbox{\boldmath $\tau$}}}

\def\bfphi{{\mbox{\boldmath $\phi$}}}

\def\bfGamma{{\mbox{\boldmath $\Gamma$}}}

\def\diag{\mathop{\rm diag}}

\def\sat{\mathop{\rm sat}}

\def\smallbfW{{\raise1.5pt\hbox{\mbox{\boldmath $_W$}}}}
\def\t{^{\rm T}}
\let\ts=\thinspace

\def\Re{\hbox{I}\!\hbox{R}}

\def\mypsfrag#1#2#3#4#5{
        \begin{figure}[htp]
           \begin{center}
              {\leavevmode
                 {\includegraphics[width=#1truecm]{#2.eps}}
              }
           \end{center}
           \vspace{#3}
           \caption{#4}
           \label{#5}
        \end{figure}
}

\def\my4psfrag#1#2#3#4#5#6#7#8{
        \begin{figure}[htp]
        \begin{center}
            \begin{tabular}[h]{c c}
              {\leavevmode{\includegraphics[width=#1truecm]{#2.eps}}}
              &
              {\leavevmode{\includegraphics[width=#1truecm]{#3.eps}}} \\
              {\leavevmode{\includegraphics[width=#1truecm]{#4.eps}}}
              &
              {\leavevmode{\includegraphics[width=#1truecm]{#5.eps}}}
         \end{tabular}
           \vspace{#6}
           \caption{#7}
           \label{#8}
        \end{center}
        \end{figure}
}

\def\mydouble4psfrag#1#2#3#4#5#6#7#8{
        \begin{figure*}[htp]
        \begin{center}
            \begin{tabular}[h]{c c}
              {\leavevmode{\includegraphics[width=#1truecm]{#2.eps}}}
              &
              {\leavevmode{\includegraphics[width=#1truecm]{#3.eps}}} \\
              {\leavevmode{\includegraphics[width=#1truecm]{#4.eps}}}
              &
              {\leavevmode{\includegraphics[width=#1truecm]{#5.eps}}}
         \end{tabular}
           \vspace{#6}
           \caption{#7}
           \label{#8}
        \end{center}
        \end{figure*}
}

\def\mymatrix#1{\left[\begin{matrix}#1\end{matrix}\right]}

\def\diag{\mathrm{diag}}

\def\zero{\hbox{\bf 0}}
\def\Zero{{\mbox{\boldmath $O$}}}

\def\bff{{\mbox{\boldmath $f$}}}
\def\bfg{{\mbox{\boldmath $g$}}}
\def\bfh{{\mbox{\boldmath $h$}}}

\def\bfp{{\mbox{\boldmath $p$}}}
\def\bfq{{\mbox{\boldmath $q$}}}

\def\bfx{{\mbox{\boldmath $x$}}}
\def\bfy{{\mbox{\boldmath $y$}}}

\def\bfC{{\mbox{\boldmath $C$}}}
\def\bfD{{\mbox{\boldmath $D$}}}

\def\bfF{{\mbox{\boldmath $F$}}}

\def\bfI{{\mbox{\boldmath $I$}}}
\def\bfJ{{\mbox{\boldmath $J$}}}
\def\bfK{{\mbox{\boldmath $K$}}}

\def\bfM{{\mbox{\boldmath $M$}}}

\def\bfGamma{{\mbox{\boldmath $\Gamma$}}}

\def\bfbeta{{\mbox{\boldmath $\beta$}}}
\def\bfgamma{{\mbox{\boldmath $\gamma$}}}

\def\bfsigma{{\mbox{\boldmath $\sigma$}}}
\def\bftau{{\mbox{\boldmath $\tau$}}}

\def\bfphi{{\mbox{\boldmath $\phi$}}}

\usepackage{accents}
\usepackage{enumitem}

\newtheorem{assumption}{Assumption}
\newtheorem{remark}{Remark}
\newtheorem{theorem}{Theorem}
\newtheorem{problem}{Problem}

\newtheorem{lemma}[theorem]{Lemma}
\newenvironment{proof}
{\textit{Proof.} }
{\hfill $\blacksquare$\\}

\def\iacdag#1{\bfJ_{\sigma}^\dag}

\renewcommand{\baselinestretch}{1.}
\usepackage{appendix}
\usepackage{graphicx}      
\usepackage{natbib}        

\begin{document}
\begin{frontmatter}
\title{Safety in human-multi robot collaborative scenarios:  a trajectory scaling approach}
\author{Martina Lippi, Alessandro Marino}
\address{University of Salerno, Via Giovanni Paolo II, 132, 84084, Salerno (SA), Italy ( email: \{mlippi,almarino\}@unisa.it)}
\thanks[footnoteinfo]{Authors are in alphabetical order}
\thanks[footnoteinfo]{The research leading to these results has received funding from the European
Community’s H2020 Program under grant agreement
n. 785419 (EU.3.4.5.4. - ITD Airframe project LABOR - Lean robotized AssemBly and cOntrol of composite aeRostructures).}
  
\begin{abstract}

In this paper, a strategy to handle the human safety in a multi-robot scenario is devised.
In the presented framework, it is foreseen that 
 robots  are in charge of performing any cooperative manipulation task which is parameterized by a proper task function.
The devised architecture answers to the increasing demand of strict cooperation between humans and robots, since it equips 
a general multi-robot cell with the feature of  making robots and human working together.
The human safety is properly handled by defining a safety index which depends both on the relative position and velocity of the human operator 
and robots. Then, the multi-robot task trajectory is properly scaled in order to ensure that the human safety never falls below a given threshold
which can be set in worst conditions according to a minimum allowed distance.
Simulations results are presented in order to prove the effectiveness of the approach.
\end{abstract}
\end{frontmatter}

\section{Introduction}\label{sec:intro}
The close cooperation of humans and robots is a highly desirable feature since it allows to benefit of the outperforming reasoning  capabilities of humans and the extreme precision and strength of robots. 
However, it is straightforward to recognize that the human safety is of the utmost importance in such a scenario which requires, at least, the robots to be controlled in such a way to not harm human operators~\cite{Haddadin_IJRR2009}.
In this regard, initial regulations about human safety with respect to industrial robots can be found in the American {ANSI/RIA R15.06}, in the {European EN 775} or in the more general international standard  {ISO 10218} and the technical specification document {ISO/TS~15066}.
In detail, the latter exactly focuses on human-robot collaborative scenarios and envisages four possible safe interactions: 
\begin{enumerate}
\item safety-rated monitored stop, i.e. robots are required to stop when humans enter the working area;
\item hand guiding, i.e. robots are required to follow human manual guidance;
\item \label{en:iso_ssm} speed and separation monitoring, i.e. robots have to keep a minimum safety distance from operators;
\item \label{en:iso_pfl} power and force limiting, i.e. robots are required to mitigate human harm in the case of impact.  
\end{enumerate}
It is clear that interactions~\ref{en:iso_ssm} and \ref{en:iso_pfl} involve integrating the robot autonomous task with human safety requirements. As highlighted in~\cite{Gomez_IEEAccess2017}, this also requires the inclusion of different sensors whose features depend on the nature of the interaction: 
from  sensors for detecting the presence of human operators for collision prevention, e.g. motion capture systems, range sensors or artificial vision systems as in~\cite{Flacco_ICRA2012}, to sensors for assessing force exchange when an impact occurs, e.g. force or tactile sensors as in~\cite{Cirillo_RAL2016}. 
Although power and force limiting is crucial in the case of physical human-robot interaction where contact is unavoidable, distance monitoring would be more suitable for pure coexistence in the working area. 
In the latter scenario, it becomes relevant to quantify the level of human safety, looking at the overall structure of the manipulator as a source of danger to humans, so that the robots behaviour can be adapted accordingly. 
An index based on distance, velocity and inertial contributions is proposed in~\cite{Kulic_RAS2006} and is evaluated for the nearest point between each link and the operator; then, such danger index is exploited to generate a virtual repulsive force according to artificial  potential field theory in~\cite{Khatib_ICRA1985}. The study presented in~\cite{Zanchettin_TRO2013} also focuses on defining an assessment of human safety which is now based on velocity and distance terms and is extended to the overall structure of the manipulator by a proper integration along each link; a gradient-based technique is then adopted to drive the manipulator. 
The previous approaches rely on pursuing evasive actions to increase safety, however, in industrial settings, it is generally recommended to follow the desired task path without deviating from it. 
This guideline is broadened in~\cite{Zanchettin_TASE2016} where robot velocity is modulated in accordance with the distance from the operator while preserving the nominal task path. A further approach is presented in~\cite{Liu_ICRA2016} and~\cite{Kimmel_TRO2017} where the safe interaction problem is modeled as an invariance control problem, i.e. it is based on defining a safe set of robot states for which no collision occurs and then on making this set invariant.
\\The previous works show how research regarding human-robot interaction and, in particular, human safety is a hot topic; however, to the best of authors' knowledge, the case of interaction between human operators and strictly cooperative robot systems has not yet been investigated. In this context, human safety must certainly remain the highest priority task, but at the same time the coordination of the robot team must be managed.
\\Motivated by these considerations, this study  presents a general solution for handling the human safety in a scenario composed by multiple cooperative robots.
Starting from the definition of a  safety index depending on the human operator's state and on the state of a generic point of the robot structure, first the safety associated to the whole robot and, then, to whole team are computed.
This safety measure is adopted to properly modify the robots trajectories in order to preserve the cooperative task and so as to not violate the safety requirements. However, because of the constraint represented by the task itself, this might result in a too restrictive strategy that might lead to the violation of the established safety requirements. If this case occurs, the task is interrupted and an impedance-based strategy is adopted; the task is then recovered when the safety conditions are restored.\\
The devised solution presents several desirable features with respect to other solutions cited above: (i) it works for general expressions of the safety index, (ii) it explicitly takes into account the multi-robot nature of the task, (iii) it does not modify the task path or require the task to be aborted unless if strictly necessary.
\\The paper is organized as follows.
Section~\ref{sec:mathbackground} introduces the mathematical background and the problem setting.
In Section~\ref{sec:humansafetyassess}, the adopted safety index is defined and analyzed in detail, while in Section~\ref{sec:humanavoidance} this index is exploited to define a  safe human-robot interaction strategy.
Finally, numerical simulations and conclusions are presented in Sections~\ref{sec:case_study} and \ref{sec:conclusions}, respectively.

\section{Mathematical Background}\label{sec:mathbackground}
In this paper, we consider a multi-robot
work-cell in which human and robots are allowed to share the same area. 
In particular, the cell is composed by
$N$~\emph{worker} robots  which 
execute the main work the cell is aimed to.
\\We assume that robots are  manipulators eventually mounted on a mobile base whose general model is
\begin{equation}\label{eq:DymRobot}
\bfM_i(\bfq_i)\ddot\bfq_i\!+\!\bfC_i(\bfq_i,\dot\bfq_i)\dot\bfq_i\!+\!\bfF_i\dot\bfq_i\!+\!\bfg_i(\bfq_i)\!=\!\bftau_i\!-\!\bfJ_i\t(\bfq_i)\bfh_i
\end{equation}
where $\bfq_i\in\Re^{n_i} $ ($\dot\bfq_i$, $\ddot\bfq_i$) is the joint position (velocity,
acceleration) vector, $\bftau_i\in\Re^{n_i}$ is the joint torque vector, $\bfM_i(\bfq_i)\in\Re^{n_i\times n_i}$
is the symmetric positive definite inertia matrix, $\bfC_i(\bfq_i,\dot\bfq_i)\in\Re^{n_i\times n_i}$ is
the centrifugal and Coriolis terms matrix, $\bfF_i\in\Re^{n_i \times n_i}$ is the matrix modeling viscous friction, $\bfg_i(\bfq_i)\in\Re^{n_i}$ is the vector of gravity terms, $\bfJ_i(\bfq_i)\in\Re^{p\times n_i}$ is the manipulator Jacobian matrix, and $\bfh_i\in\Re^{p}$ is the vector of interaction forces between the robot's end-effector and the environment. 
Let $\bfq_{r,i}(t)\in\Re^{n_i} $ ($\dot\bfq_{r,i}(t)$, $\ddot\bfq_{r,i}(t)$) be the joint position (velocity,
acceleration) reference of the $i$\ts th robot, the following assumption is made.
\begin{assumption}\label{ass:innermotionloop}
Each robot is equipped with an inner motion control loop which guarantees tracking of a reference joint trajectory, i.e. $\bfq_{r,i}\approx\bfq_i$ ($\dot \bfq_{r,i}\approx\dot\bfq_i$, $\ddot\bfq_{r,i}\approx\ddot\bfq_i$).
\end{assumption}
This assumption is realistic for all commercial platforms and makes the devised solution suitable also for off-the-shelf robotic platforms for which   the low level control layer is generally not made accessible to directly set the $\bftau_i$ input in \eqref{eq:DymRobot}.
\\The second order kinematic relationship is such as
\begin{equation}\label{eq:kinemodel}
 \ddot\bfx_i=\bfJ_i(\bfq_i)\ddot\bfq_i+\dot{\bfJ}(\bfq_i)\dot\bfq_i=\bfJ_i(\bfq_i)\bfy_i+\dot{\bfJ}(\bfq_i)\dot\bfq_i
\end{equation}
where  $\bfx_i=\mymatrix{\bfp_i\t, \bfphi_i\t}\t\in\Re^{p}$ is the end-effector configuration of the $i$\ts th manipulator with respect to the world frame expressed in terms of position $\bfp_i$ and orientation $\bfphi_i$,
 and $\bfy_i=\ddot\bfq_i$ is the input of the assumed virtual model. For the sake of notation compactness, the dependence of $\bfJ_i$ from its parameter $\bfq_i$ is generally omitted in the following.
  \\For the purpose of the overall description of the cell, let us introduce the collective vectors
  \begin{equation}
  \begin{aligned}
\bfx&=\mymatrix{\bfx_1\t, & \bfx_2\t, & \ldots,& \bfx_N\t}\t\in\Re^{Np}\\
\bfq&=\mymatrix{\bfq_1\t, & \bfq_2\t, & \ldots,& \bfq_N\t}\t\in\Re^{n}\\
\bfJ(\bfq)&=\diag\{\bfJ_1(\bfq_1),\dots, \bfJ_N(\bfq_N)\}\in\Re^{Np\times n}
  \end{aligned}
  \end{equation}
where $n=\displaystyle \sum_{i}n_i$.
\\In what follows, with
${\Zero}_{m}$ and ${\bfI}_{m}$ we denote the null and identity matrices in $\Re^{m\times m}$, respectively, and with ${\bf0}_{m}$ we denote the column vector in $\Re^{m}$ with all zero elements.

\subsection{Problem setting}\label{ssec:task}
It is assumed that the cooperative task assigned to robots is defined by means of a task function 
 $\bfsigma=\bfsigma(\bfx)\in\Re^m$ as
\begin{equation}\label{eq:task}
\bfsigma = \bfJ_{\sigma}\bfx,\,\,\,\,\dot\bfsigma = \bfJ_{\sigma}\dot\bfx,\,\,\,\,\ddot\bfsigma = \bfJ_{\sigma}\ddot\bfx 
\end{equation}
being $\bfJ_\sigma\in\Re^{m\times Np}$ the task Jacobian matrix.
A flexible formulation for the task function $\bfsigma$ is given by the absolute-relative variables as in~\cite{caccavale12}. 
In detail, the absolute variables define the position and orientation of the centroid of the end-effector configurations: 
\begin{equation}\label{eq:sig_abs}
\bfsigma_1=\frac{1}{N}\sum_{i=1}^N \bfx_i=\bfJ_{\sigma_1} \bfx
\end{equation} 
with $\bfJ_{\sigma_1}=\frac{1}{N} {\bf{1}}_N\t \otimes \bfI_p \in \Re^{p \times Np}$, while the relative variables represent the team formation: 
\begin{equation}\label{eq:sig_rel}
\bfsigma_2=[(\bfx_N-\bfx_{N-1})\t \ldots (\bfx_2-\bfx_{1})\t]\t=\bfJ_{\sigma_2} \bfx
\end{equation} 
with 
\begin{equation}\label{eq:jac_sig_rel}
\bfJ_{\sigma_2}=\mymatrix{-\bfI_p & \phantom{-}\bfI_p & \Zero_p & \ldots & \Zero_p \cr \Zero_p  & -\bfI_p & \bfI_p  & \ldots & \Zero_p \cr \vdots & & \ddots &  & \vdots \cr \Zero_p & \ldots & \Zero_p & -\bfI_p & \bfI_p } \in \Re^{(N-1)p \times Np}
\end{equation}
Hence, in virtue of~\eqref{eq:sig_abs} and \eqref{eq:sig_rel}, the task function in~\eqref{eq:task} is
\begin{equation}\label{eq:task_tot}
\bfsigma = \mymatrix{\bfsigma_1 \cr \bfsigma_2} = \mymatrix{\bfJ_{\sigma_1} \cr \bfJ_{\sigma_2}}\bfx = \bfJ_\sigma \bfx
\end{equation}
with $\bfJ_\sigma \in \Re^{Np \times Np}$ and $m=Np$.

The  objective is to compute the 
input $\bfy_i$
in~\eqref{eq:kinemodel} in order to have $\bfsigma$ tracking a nominal task trajectory $\bfsigma_n(t)$, allowing human operators to enter the cell during execution.
In such a scenario,  the safety of the humans is the highest priority task and the robot trajectory must be modified accordingly.
To the aim, the nominal trajectory $\bfsigma_n(t)$ is first properly modified in order to generate a human-safe trajectory $\bfsigma_r(t)$ which is the trajectory actually tracked as it will be detailed in the following.
\\Moreover, herein it is not of interest to design algorithms for human detection, while the focus is on defining a human-safe strategy for the coordination of cooperating robots. 
Hence, the following assumption is made. 
\begin{assumption}\label{ass:humanestimation}
If human operators are in the nearby of the work-cell, either robots are able to detect them or this information is made available to robots.
This information might concern, for instance, the position of the head or the chest of the human, or a set of representative points.
\end{assumption}
As stated above, the control input for the $i$\ts th robot has to be such that, globally, the cooperative task 
described according to the task function in~\eqref{eq:task_tot} 
tracks the reference $\bfsigma_r(t)$.
Hence, the $i$\ts th virtual input in~\eqref{eq:kinemodel} can be selected  resorting to a standard closed loop inverse kinematic law:
\begin{equation}\label{eq:jointinput}
\begin{aligned}
\bfy_i \!=& \ddot\bfq_i=\bfJ_i^\dag\!\left(\bfGamma_i\bfJ_\sigma^\dag\!\left(\ddot \bfsigma_r\!+\!k_\sigma\dot{\tilde{\bfsigma}} +\lambda_\sigma{\tilde{\bfsigma}}\right) \!-\!\dot\bfJ_i\dot\bfq_i \!\right) +\ddot{\bfq}_{n,i} \\
\end{aligned}
\end{equation}
being 
${\tilde {\bfsigma}}=(\bfsigma_r-{ {\bfsigma}})\in\Re^m$ the task tracking error, $\ddot{\bfq}_{n,i}\in\Re^{n_i}$ an arbitrary vector of joint accelerations such as $\bfJ_i(\bfq_i)\ddot{\bfq}_{n,i}=\zero_p$ which might be exploited to locally optimize secondary objectives, 
$k_\sigma$, $\lambda_\sigma$ positive gains and
\begin{equation}
	\bfGamma_i = \{\begin{matrix}\vspace{-5mm}\Zero_{p} & \cdots & \underbrace{\bfI_{p}}_{i\hbox{~th robot}}
	& \cdots & \Zero_{p}\end{matrix}\}\in\Re^{p \times Np}
\end{equation}
\vspace{0.2cm}\\ 
a selection matrix.
It is easy to recognize that in virtue of~\eqref{eq:task} and \eqref{eq:kinemodel} it holds
$$
\bfJ_\sigma(\bfJ\bfy+\dot\bfJ\dot\bfq)=\bfJ_\sigma\ddot\bfx=\ddot\bfsigma=\ddot \bfsigma_r\!+\!k_\sigma\dot{\tilde{\bfsigma}} +\lambda_\sigma{\tilde{\bfsigma}}
$$
where $\bfy=\mymatrix{\bfy_1\t,\dots,\bfy_N\t}\t$ and which leads to the following exponentially stable linear second order dynamics
$$
\ddot{\tilde{\bfsigma}}+\!k_\sigma\dot{\tilde{\bfsigma}} +\lambda_\sigma{\tilde{\bfsigma}}={\zero}_m
$$

\section{Human safety assessment}\label{sec:humansafetyassess}
In this section, we focus on formulating an index to assess the level of human safety with respect to the team of robots. 
The devised safety strategy tries to have the
robots follow the task trajectory $\bfsigma_n(t)$  as much as possible in compliance with human safety requirements. 
The basic idea 
is to parameterize  the  nominal trajectory for $\bfsigma$ in 
\eqref{eq:task_tot} through
a non-negative non-decreasing scalar function 
$${s_n: t\in[t_0\, t_f] \rightarrow \Re}$$ 
with $t_0$ and $t_f$ the initial and final time instant, respectively, and to have the robots cooperatively track 
\begin{equation}\label{eq:sr}
\bfsigma_r(t)=\bfsigma_n(s_r(t))
\end{equation}
where ${s_r: t\in\Re \rightarrow \Re}$ is a properly {\it scaled} version of $s_n(t)$ which takes into account the human safety; obviously, this strategy allows the robots to preserve the task {path}.
\\Let us introduce a general 
 safety index which allows to quantify the level of human safety with respect to a generic moving point $P$ belonging to the robot structure 
\begin{equation}\label{eq:f}
f(\bfp,\dot \bfp,\bfp_o,\dot \bfp_o ) = \alpha_1(d)+\alpha_2(d,\dot d) 
\end{equation}
where  $\bfp\in\Re^3$ and $\dot \bfp\in\Re^3$ are the position and velocity of point $P$, respectively,  $d=\|\bfp-\bfp_o\|$ is the distance between the point  and the human operator's position $\bfp_o\in\Re^3$ assumed to be available (see Assumption~\ref{ass:humanestimation}), $\dot d$ is the distance derivative and $\alpha_1$, $\alpha_2$ are generic scalar functions such as the following properties hold:
\begin{enumerate}[label=Property \arabic*.,itemindent=*]\label{en:cond_alpha}
\item $\alpha_1(d)$ is a non negative continuous monotonically increasing function with respect to $d$;
\item $\alpha_2(d,\dot d) $ is a continuous monotonically increasing function with respect to  $\dot d$ and such that:
\begin{enumerate}
\item  {$\displaystyle\lim_{\dot d \rightarrow +\infty} \alpha_2(d,\dot d)=c$, $\forall d$}  with $c\in\Re^+$; 
\item { $\frac{\partial \alpha_2(d,\dot d)}{\partial \dot d} \neq 0$}  $\forall d$ and $\forall \dot d\neq \infty$. 
\end{enumerate}
\end{enumerate}
The ratio behind Property $1$ is that  the human-safety with respect the point $P$ increases with the distance $d$.
Concerning Property $2$,  function $\alpha_2$ is such as the safety index increases for positive values of $\dot d$ with a slope that might be modulated by $d$. The motivation behind the asymptotic bound $c$ in Property $2$(a)  for $\dot d\rightarrow +\infty$ is that it prevents the safety index to reach a too high value for high values of $\dot d$ and arbitrarily small values of the distance $d$; in this way, the distance parameter is always the high priority feature.
Finally, Property $2$(b)  ensures that for finite values of $\dot d$ the index $f$ is sensitive to variation of velocity $\dot d$ such as by changing $\dot d$ the value of $f$ can be modified.

By leveraging the approach in~\cite{Zanchettin_TRO2013}, the evaluation of the safety function in~\eqref{eq:f} can be easily extended 
to the entire structure of the $i$\ts th manipulator by properly integrating \eqref{eq:f} along  its structure and obtain a cumulative safety index $F^i$. 
In particular, the measure of human safety with respect to the $l$\ts th link of the  $i$\ts th manipulator can be obtained by 
integrating $f$ along the volume $\mathcal{V}_l$ of  link $l$ 
\begin{equation}\label{eq:fc_ic}
F^i_l=\int_{\mathcal{V}_l} f (\bfp,\dot \bfp,\bfp_o,\dot \bfp_o )~d\bfp
\end{equation}  
In order to make the computation of \eqref{eq:fc_ic} affordable, 
the generic link of the $i$\ts th robot is simplified as a segment starting at $\bfp^i_{l,0}$ and ending at $\bfp^i_{l,1}$, thus~\eqref{eq:fc_ic} 
becomes
\begin{equation}\label{eq:fc_is}
\left\{
\begin{aligned}
\!\!F^i_{l}&=\int_0^1 \!
f (\bfp^i_{l,r},\dot \bfp^i_{l,r},\bfp_o,\dot \bfp_o )dr\\
\bfp^i_{l,r}&=\bfp^i_{l,0}\!+r(\bfp^i_{l,1}-\bfp^i_{l,0})\\
\dot \bfp^i_{l,r}&=\dot \bfp^i_{l,0}\!+r(\dot \bfp^i_{l,1}-\dot \bfp^i_{l,0})\\
\end{aligned}
\right.
\end{equation}
Finally, the safety index associated to the   $i$\ts th manipulator with $n_{l}^i$ links is
\begin{equation}\label{eq:fc_i}
\!\!F^i\!\!=\displaystyle \sum_{l=1}^{n^i_{l}+1} F^i_{l}
\end{equation}
where an additional virtual link is introduced to account for the end-effector and the cooperative task to achieve. 
Concerning the derivative of the safety measure in~\eqref{eq:fc_i}, the following lemma holds.
\begin{lemma}
The derivative of the cumulative safety function~\eqref{eq:fc_i} associated to the $i$\ts th robot  is linear in the path parameter acceleration $\ddot s_r(t)$, i.e. it is
\begin{equation}\label{eq:dfc_i}
\begin{aligned}
\dot F^i &= \mu^i_{1} \,\ddot s_r  + \mu^i_{2}
\end{aligned}
\end{equation}
where the expressions of $\mu^i_{1}$, $\mu^i_{2}\!\!\in\!\Re$ are provided in the proof.
\end{lemma}

\noindent
\begin{proof}
Let us consider the time derivative of the safety function in~\eqref{eq:f} associated with a generic point {$\bfp^i_{l,r}$ ($r\in[0\, 1]$)} belonging to the $l$\ts th link of the $i$\ts th robot; it is
\begin{equation}\label{eq:dfj}
\begin{aligned}
 \dot f\!\!
 &=\!\! \left(\frac{\partial \alpha_1(d^i_{l,r})}{\partial d^i_{l,r}} +\frac{\partial \alpha_2(d^i_{l,r},\dot d^i_{l,r})}{\partial d^i_{l,r}}\right)\dot d^i_{l,r}+\frac{\partial \alpha_2(d^i_{l,r},\dot d^i_{l,r})}{\partial \dot d^i_{l,r}} \ddot d^i_{l,r}
\end{aligned}
\end{equation}
with $d^i_{l,r}=\|\bfp^i_{l,r}-\bfp_o\|$,
whose second time derivative  is
\begin{equation}\label{eq:dd_dj}
\ddot d^i_{l,r}= \bfbeta_1\t \ddot \bfp^i_{l,r} + \beta_2
\end{equation}
where coefficients $\bfbeta_1\in\Re^3$, $\beta_2\in\Re$ are defined as  
\begin{equation*}
\left\{
\begin{aligned}
\bfbeta_1&=\frac{\bfp^i_{l,r}-\bfp_o}{d^i_{l,r}} & d^i_{l,r}\neq 0\\
\beta_2&=\!-\bfbeta_1\t\ddot \bfp_o\!+\!\frac{\|\dot \bfp^i_{l,r}\!-\!\dot \bfp_o\|}{d^i_{l,r}}^2\!-\!\frac{[\bfbeta_1\t(\dot \bfp^i_{l,r}\!-\!\dot \bfp_o)]}{d^i_{l,r}}^2 & d^i_{l,r}\neq 0
\end{aligned}
\right.
\end{equation*}
\\At this point, let us consider the well-known relation between the linear acceleration of a point belonging to the structure of a manipulator and its joint variables, i.e.
\begin{equation}\label{eq:ddpj_kin}
\ddot \bfp^i_{l,r}=\bfJ^i_{l,r} (\bfq_i) \ddot \bfq_i+\dot \bfJ^i_{l,r}(\bfq_i, \dot \bfq_i) \dot \bfq_i
\end{equation}
being $\bfJ_{l,r}^i\in\Re^{3 \times n_i}$ the positional Jacobian matrix associated to  $\bfp^i_{l,r}$. 
Now, by partially deriving the reference task function in~\eqref{eq:sr} with respect to $s_r$, it holds
\begin{equation*}
\dot \bfsigma_r= \frac{\partial \bfsigma_r}{\partial s_r}\dot s_r, \quad \ddot \bfsigma_r= \frac{\partial^2 \bfsigma_r}{\partial s_r^2}\dot s_r^2+\frac{\partial \bfsigma_r}{\partial s_r}\ddot s_r 
\end{equation*}
and in virtue of~\eqref{eq:jointinput}, equation~\eqref{eq:ddpj_kin} can be expressed as
\begin{equation}\label{eq:ddpj}
\ddot \bfp^i_{l,r}=\bfgamma_1 \ddot s_r + \bfgamma_2
\end{equation}
where $\bfgamma_1$,$\bfgamma_2\in\Re^3$ are defined as
\begin{equation*}
\!\!\!\!\left\{
\begin{aligned}
\!\bfgamma_1&=\bfJ^i_{l,r}\bfJ_{i}^\dagger \bfGamma_{i}\iacdag{i}\frac{\partial \bfsigma_r}{\partial s_r} \\
\!\bfgamma_2&=\bfJ^i_{l,r}[\bfJ_i^\dag\bfGamma_i\bfJ_\sigma^\dag\!\left(\frac{\partial^2 \bfsigma_r}{\partial s_r^2}\dot s_r^2+\!k_\sigma\dot{\tilde{\bfsigma}} +\lambda_\sigma{\tilde{\bfsigma}}\right) \!\!-\!\bfJ_i^\dag\dot\bfJ_i\dot\bfq_i\!+\!\ddot{\bfq}_{n,i}]\\
& +\dot \bfJ^i_{l,r}\dot \bfq_i
\end{aligned}
\right.
\end{equation*}
By folding \eqref{eq:dd_dj} and \eqref{eq:ddpj} in \eqref{eq:dfj}, the term $\dot f$ becomes
\begin{equation}\label{eq:df_point}
\begin{aligned}
\dot f &= \lambda_1 \ddot s_r  + \lambda_2
\end{aligned}
\end{equation}
where the expressions of $\lambda_1$ and $\lambda_2\in\Re$ are
\begin{equation*}
\left\{
\begin{aligned}
\lambda_1 &=(\bfbeta_1\t\bfgamma_1)\frac{\partial \alpha_2}{\partial \dot d^i_{l,r}} \\ 
\lambda_2 &=(\bfbeta_1\t\bfgamma_2 + \beta_2)\frac{\partial \alpha_2}{\partial \dot d^i_{l,r}}+\left(\frac{\partial \alpha_1}{\partial d^i_{l,r}} +\frac{\partial \alpha_2}{\partial d^i_{l,r}}\right)\dot d^i_{l,r}
\end{aligned}
\right.
\end{equation*}
Therefore, in virtue of~\eqref{eq:fc_is},\eqref{eq:fc_i} and \eqref{eq:df_point}, it finally holds
\begin{equation*}
\dot F^i =  \mu^i_{1} \,\ddot s_r  + \mu^i_{2}
\end{equation*}
with $ \mu^i_{1}$, $ \mu^i_{2}\in\Re$ defined as below
\begin{equation*}
\left\{
\begin{aligned}
\mu_{1}^i&= \sum_{j=1}^{n^i_{l}+1} \int_0^1 \lambda_1(\bfp^i_{l,r}, \dot \bfp^i_{l,r}, \bfp_o, \dot \bfp_o, \bfq_i, \dot \bfq_i, s_r)~dr  \\
\mu_{2}^i&=  \sum_{j=1}^{n_{l}^i+1} \int_0^1 \lambda_2(\bfp^i_{l,r}, \dot \bfp^i_{l,r}, \bfp_o, \dot \bfp_o, \ddot \bfp_o, \bfq_i, \dot \bfq_i, \ddot \bfq_{n,i}, s_r, \dot s_r)~dr 
\end{aligned}
\right.
\end{equation*}
where the dependencies of $\lambda_1$ and $\lambda_2$ on their parameters are now made explicit for the sake of completeness.\\
This completes the proof.
\end{proof}

In the multi-robot case, the overall safety function $F$  (and its derivative), which accounts for all the worker robots in the team, can be easily deduced by combining the  safety functions in \eqref{eq:fc_i} associated to each manipulator 
\begin{equation}\label{eq:fc}
F=\sum_{i=1}^N F^i, \quad \dot F=\sum_{i=1}^N \dot F^i
\end{equation}
We are now ready to formally state the following problem.

\begin{problem}\label{pr:problem}
Let us consider a multi-robot system composed by $N$ mobile manipulators performing the cooperative task defined as in~\eqref{eq:task_tot} for which a desired trajectory $\bfsigma_n(s_n(t))$ parametrized with respect to a scalar function $s_n(t)$ is assigned. Moreover, let us also assume that a minimum  value $F_{min}$ for function $F$ in~\eqref{eq:fc} is assigned; then, our objective is to properly scale $\bfsigma_n(t)=\bfsigma_n(s_n(t))$ so as to generate a new reference trajectory $\bfsigma_r(t)=\bfsigma_n(s_r(t))$ such that $F\ge F_{min}$, $\forall t$.
\end{problem}
\begin{remark}\label{rm:fmin}
Problem~\ref{pr:problem} requires that a  minimum value $F_{min}$ for function $F$ is defined; thus, the problem arises on how to choose this lower bound.
A first strategy consists in tuning $F_{min}$ via experimental trials based on the human feeling about the {\it experienced}  level of safety resorting to techniques similar to~\cite{RajendraAcharya2006}.
Another strategy consists in selecting $F_{min}$  such as, defined  $d$ as the distance between the human operator and the team
\begin{equation}\label{eq:dmin}
d=\displaystyle \min_{\begin{aligned}\forall i, l,r  \end{aligned}}\|\bfp^i_{l,r}-\bfp_o\|
\end{equation}
$F \ge F_{min}$ ensures that $d\ge d_{min}$ for some value ${d_{min}>0}$~(\cite{Zanchettin_TRO2013}).
As an example, the second strategy is pursued in the Section~\ref{sec:case_study}.
\end{remark}
The next section provides a possible solution to Problem~\ref{pr:problem}.
\section{The human-robot  avoidance strategy}\label{sec:humanavoidance}

An overview of the devised strategy for solving Problem~\ref{pr:problem} is provided in Figure~\ref{fig:planner_scheme}; in particular, the proposed approach foresees tracking the nominal trajectory until the level of human safety is above the minimum accepted value, i.e. $F>F_{min}$; if the nominal trajectory leads the safety level to its minimum value, then a velocity modulation is applied while preserving the nominal path and, if this is not enough to guarantee $F\geq F_{min}$, then the requirement of preserving the path is relaxed.
It is worth remarking that, in the case of redundant robots, each manipulator also exploits the extra degrees of freedom to maximize the cumulative safety index. 
\vspace{-0.1cm}
\begin{psfrags}
\def\scal{0.8}  
\def\scalc{0.7} 
\def\scalt{0.7} 
 \psfrag{b1}[cc][][\scal]{\begin{tabular}{@{}c@{}}Nominal trajectory \\tracking ($F>F_{min}$)\end{tabular} }
\psfrag{b2}[cc][][\scal]{\shortstack[c]{Scaled trajectory \\tracking}}
\psfrag{b3}[cc][][\scal]{ Path deformation }
\psfrag{l1}[lc][][\scalt]{\hspace{-0.6cm}\shortstack[c]{Trajectory scaling \\ ($F = F_{min}$)} }
\psfrag{l2}[lc][][\scalt]{\hspace{-0.3cm}\shortstack[c]{Recovery of the \\ nominal trajectory} }
\psfrag{l3}[lc][][\scalt]{\hspace{-0.6cm}  \shortstack[c]{Constraint violation \\ ($F<F_{min}$)}}
\psfrag{l4}[cc][][\scalt]{\hspace{-0.7cm}\shortstack[c]{Zero deformation \\terms \\($\Delta \bfp,\,\Delta \dot \bfp,\,\Delta \ddot \bfp $)} }
\psfrag{l4}[cc][][\scalt]{\hspace{-0.9cm}\shortstack[c]{Recovery of the \\ nominal trajectory}}
\mypsfrag{7}{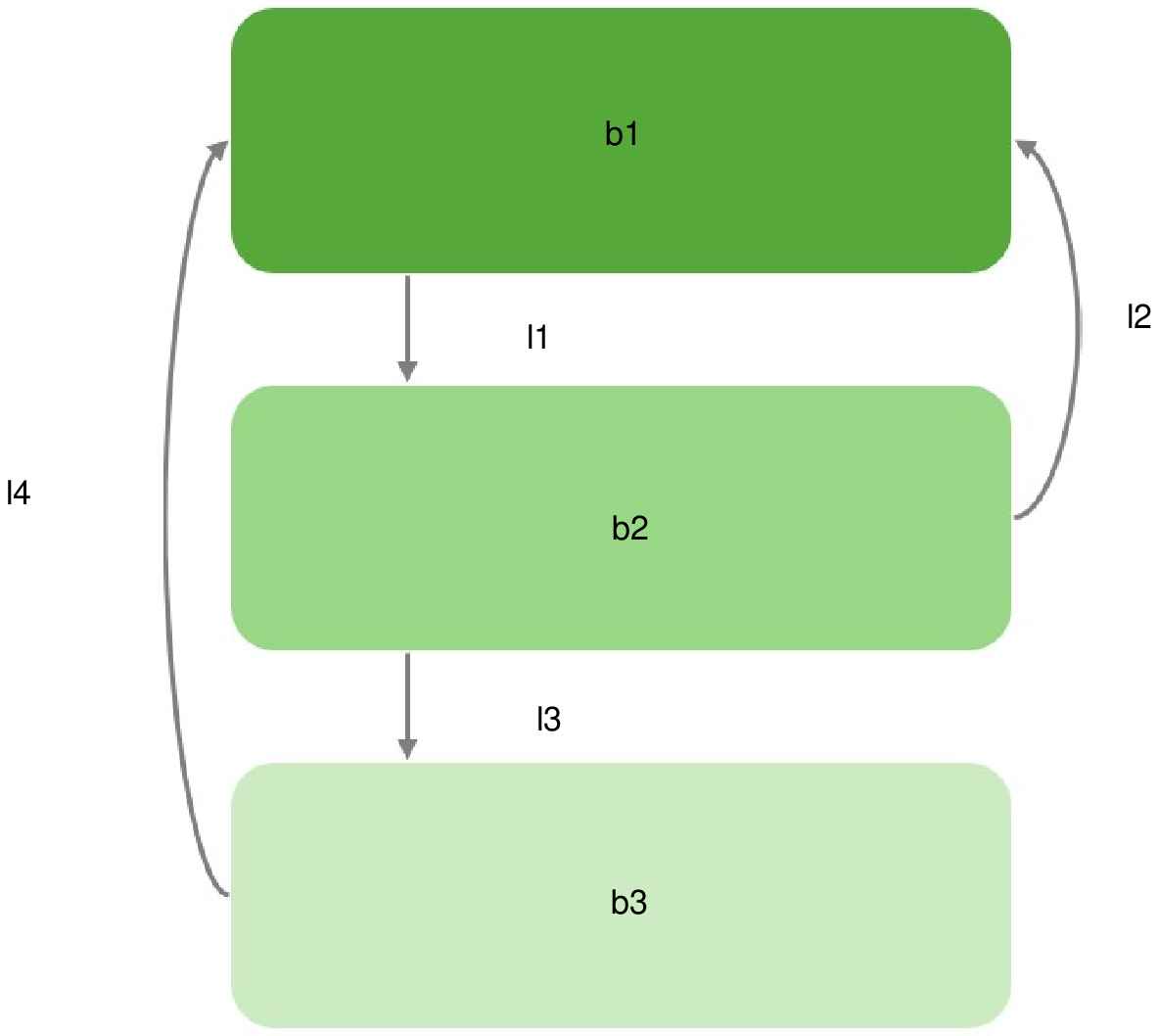}{-28pt}{High-level scheme of the human avoidance strategy; transition conditions are detailed in Section~\ref{sec:humanavoidance}. }{fig:planner_scheme} 
\end{psfrags}

\subsection{Human-robot avoidance via trajectory scaling}
By leveraging the approach in~\cite{Dahl90} designed for torque-limited path following of industrial robots,
a scaling parameter $s_r(t)$ is introduced that is function of $s_n(t)$ according to the following relation
\begin{equation}\label{ScalingFactor}
\left\{
\begin{aligned}
s_r(t) &= s_n(t)+{\Delta s}(t)\\
\dot{s}_r(t) & = \dot s_n(t)+\dot{\Delta s} (t)\\
\ddot{s}_r(t) & = \ddot s_n(t) + \ddot{\Delta s} (t)\\
\end{aligned}
\right.
\end{equation}
where $\Delta s(t)$ ($\dot{\Delta s},\ddot{\Delta s}$) might be either negative or positive and is adopted to properly scale the nominal path parameter while it is such as $\Delta s(t)=\dot\Delta s(t)=\ddot\Delta s(t)=0$ (i.e., $s_n(t)=s_r(t)$) in nominal conditions (no safety issue arises). 
Moreover, it is required that at any instant 
\begin{align}
 \dot{\Delta s} (t) &\geq  -\dot s_n(t) \label{eq:ScalingConditions1}\\
 s_n(t)+{\Delta s} (t) &\leq  s_n(t_f) \label{eq:ScalingConditions2}
\end{align}
The constraints~\eqref{eq:ScalingConditions1} and~\eqref{eq:ScalingConditions2} 
ensure that no reverse motion occurs along the path and that the end-point of the nominal trajectory is not overcome, respectively.\\
By folding \eqref{eq:dfc_i} in \eqref{eq:fc}, the expression of $\dot F$  can be stated as follows
\begin{equation}
\dot F=\mu_1 \ddot \Delta s + \mu_2
\end{equation}
with $\mu_1$, $\mu_2\in \Re$ defined as 
\begin{equation*}
\left\{
\begin{aligned}
\mu_1&=\sum_{i=1}^N \mu^i_{1} \\
\mu_2&=\ddot s_n\sum_{i=1}^N \mu^i_{1} + \sum_{i=1}^N \mu^i_{2}
\end{aligned}
\right.
\end{equation*}
and which is linear in $\ddot \Delta s$.
At this point, we are ready to determine the scaling terms $\Delta s$, $\dot\Delta s$ and $\ddot\Delta s$ such that the minimum safety condition is met. 
Therefore, starting from the constraint $F\ge F_{min}$, the lower ($\ddot{\Delta s}_{min}$) and the upper ($\ddot{\Delta s}_{max}$) bounds on the parameter $\ddot {\Delta s}$ are computed as
\begin{equation}\label{eq:max_bounds}
\ddot{\Delta s}_{max}=\left\{
\begin{array}{ll}
-{\mu}_2 /{\mu}_1, & {\mu}_1<0 \wedge F=F_{min}\\
+\infty, & \text{otherwise} \\
\end{array} 
\right.
\end{equation}
and
\begin{equation}\label{eq:min_bounds}
\ddot{\Delta s}_{min}=\left\{
\begin{array}{ll}
-{\mu}_2 /{\mu}_1, & {\mu}_1>0 \wedge F=F_{min}\\
-\infty, & \text{otherwise} \\
\end{array} 
 \right.
\end{equation}
The ratio behind \eqref{eq:max_bounds} and \eqref{eq:min_bounds} is that, as long as $F>F_{min}$, no constraint on $\dot F$ (and then on $\ddot{\Delta s}_{min}$ and $\ddot{\Delta s}_{max}$) is set; while, in the case $F=F_{min}$, the computed bounds are such as $\ddot{\Delta s}_{min} \leq \ddot{\Delta s} \leq \ddot{\Delta s}_{max}$ ensures  that $\dot F\ge 0$ and, then,  that $F$ does not fall below $F_{min}$. 
\\The derived bounds 
in~\eqref{eq:max_bounds} and~\eqref{eq:min_bounds} are used within the following dynamic system to compute $\ddot{\Delta s}$:
\begin{equation}\label{eq:scaling_dyn}
\left\{\begin{aligned}
 \ddot{\Delta s} &= -k_d \dot{\Delta s} - k_p {\Delta s}\\
 \ddot{\Delta s} &= \sat(\ddot{\Delta s}, \ddot{\Delta s}_{min}, \ddot{\Delta s}_{max}) 
 \end{aligned}
 \right.
\end{equation}
where ${\Delta s}(t_0)=\dot{\Delta s}(t_0)=0$, $k_d$ and $k_p$ are positive constants 
and $\sat()$ 
is any saturation function that bounds $\ddot{\Delta s}$ in the range 
$[\ddot{\Delta s}_{min}, \ddot{\Delta s}_{max}]$. 
The first equation in~\eqref{eq:scaling_dyn}  is such as to continuously bring $\Delta s$ to zero (i.e., $s_r$ to $s_n$), while, in the second equation, this value is saturated according to
the computed bounds. Thus, when the saturation function does not alter the input value, the scaling term $\Delta s$ ($\dot \Delta s$, $\ddot \Delta s$)  asymptotically converges to zero.
\begin{remark}\label{rm:scaling}
Constraints in~\eqref{eq:ScalingConditions1} and \eqref{eq:ScalingConditions2} imply that the scaling strategy does not generally guarantee the condition $F\geq F_{min}$ to be fulfilled. 
For example, in the case 
$\dot{\Delta s} = -\dot s_n $  and $\ddot\Delta s_{max}<0$, it is evident that further scaling would violate the constraint in~\eqref{eq:ScalingConditions1}.
\end{remark}

Because of Remark \ref{rm:scaling},  a different avoidance strategy is presented in the following section which is adopted when the condition $F\ge F_{min}$  cannot be secured by the scaling strategy presented above and that, in brief,  allows the path constraint to be violated (see Figure \ref{fig:planner_scheme}).

\subsection{Human-robot avoidance via nominal path deformation}\label{sssec:obj_imp}

In the case the dynamics in~\eqref{eq:scaling_dyn} leads to one of the constraints~\eqref{eq:ScalingConditions1} and~\eqref{eq:ScalingConditions2} being violated, 
the cooperative task is aborted and other avoidance strategies need to be adopted.
To this aim, two cases should be considered:
 loosely connected robots and 
 tightly connected robots (as in a multi-robot transportation task of rigid objects).
In the first case, once the task has been aborted, the human safety can be guaranteed independently by each robot adopting, for example, the approach devised in~\cite{Zanchettin_TRO2013}. Therefore, this case is not investigated in this paper.
In the more interesting case of tightly connected robots, the avoidance strategy needs to be compliant with the kinematic constraint that consists in having $\bfsigma_2$ constant in any reference frame attached to the grasped object.
For this reason, the avoidance strategy must consist in properly modifying $\bfsigma_1$ and, as in the previous case, exploiting the local redundancy.
In detail, let $t_s$ be the time instant in which the path constraint is relaxed, then 
the reference trajectory is modified as 
\begin{equation}\label{eq:pos_impedance}
\left\{
\begin{aligned}
\bfsigma_r(t) &=  \bfsigma_r(t_s^-)+\Delta\bfsigma_{r}(t)\\
\dot\bfsigma_r(t) &= \Delta\dot\bfsigma_r(t)\\
\ddot\bfsigma_r(t) &= \Delta\ddot\bfsigma_r(t) \\
\end{aligned}
\right.
\end{equation}
where the displacement $\Delta \bfsigma_r(t)$  is computed according to the following dynamics 
\begin{equation}\label{eq:delta_impedance}
\left\{
\begin{aligned}
&\bfM\ddot{\Delta\bfsigma}_{r}(t) + \bfD\dot{\Delta\bfsigma}_{r}(t) + \bfK{\Delta\bfsigma}_{r}(t) = \bff_r(t) \\
&\Delta{\bfsigma}_{r}(t_s)={\bf0}_m \\
&\dot{\Delta\bfsigma}_{r}(t_s)=\dot\bfsigma_{r}(t_s^-)
\end{aligned}
\right.
\end{equation}
with $\bfM,\,\bfD,\,\, \bfK\in\Re^{m\times m}$ positive definite matrices and $\bff_r\in\Re^m$ a  virtual force to be properly defined; the initial conditions in~\eqref{eq:delta_impedance} are such that to guarantee the continuity of the reference trajectory in the switching time $t_s$. 
Regarding the virtual force in~\eqref{eq:delta_impedance}, it is selected as 
\begin{equation}\label{eq:rep_force}
\bff_r=\mymatrix{-k_r\,f_r(F)\frac{\nabla F\t}{\|\nabla F\|} \quad\quad \zero_{m-3}\t}\t
\end{equation}
where $k_r$ is a positive gain, $\nabla F\in\Re^3$ is the gradient of the cumulative safety function with respect to $\bfp_{o}$ and $f_{r}(F)$ is a monotonically non increasing function of the safety index $F$ which converges to the origin for $F$ sufficiently high, i.e. $F>F_{min}+\Delta F$ with $\Delta F\in\Re^+$. 
Thus, the repulsive force is such to modify the reference centroid position with an intensity that increases when the safety value decreases  and is zero when  safety is restored; 
concerning the direction, it is opposite to that of the gradient $\nabla F$ since it represents the direction in which the operator should move to maximise $F$. 
A possible choice of $f_{r}(F)$ is shown in Figure~\ref{fig:fr_coeff}. 

\begin{psfrags}
\def\scal{0.8}  
\def\scallegend{0.6}  
\def\scalnum{0.6}
 \psfrag{f}[cc][][\scal]{ $F$}
 \psfrag{fr}[cc][][\scal]{ $f_r$}
 \psfrag{fmin}[cc][][\scal]{$F_{min}$}
 \psfrag{fdelta}[cc][][\scal]{$F_{min}+\Delta F$}
 \psfrag{0}[cc][][\scalnum]{0}
 \psfrag{1}[cc][][\scal]{$1$} 
\mypsfrag{6.7}{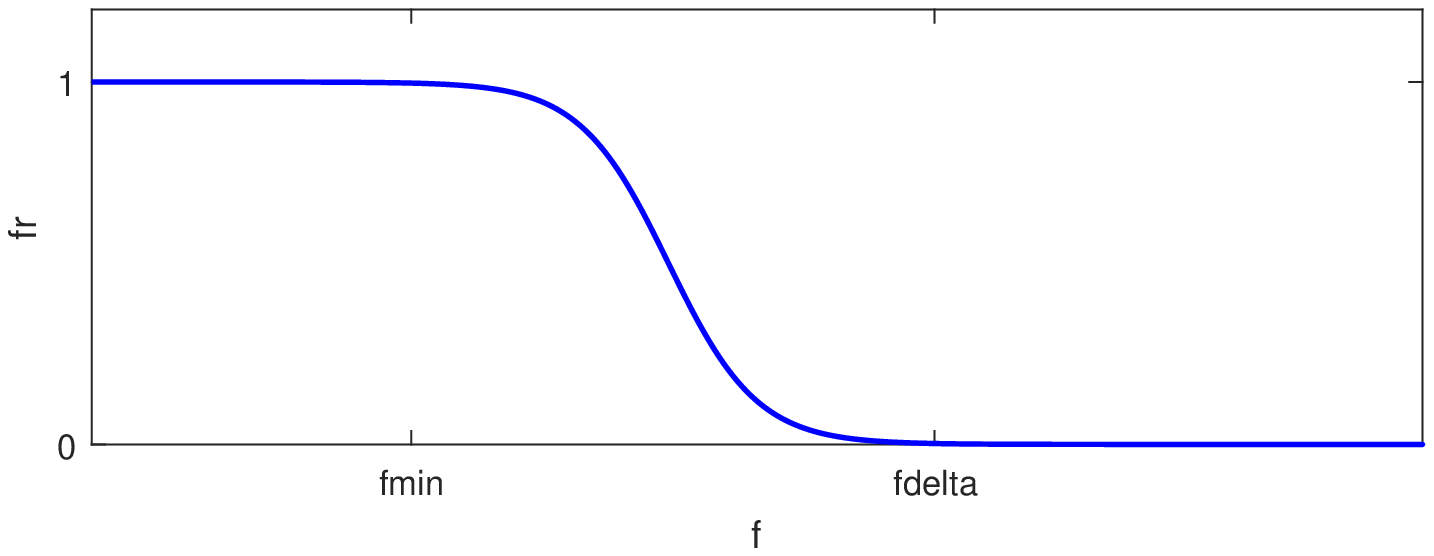}{-6pt}{Value of the intensity $f_r(F)$ in~\eqref{eq:rep_force}. }{fig:fr_coeff} 
\end{psfrags} 


Finally, when the following conditions are met: 
\begin{enumerate}
 \item $\bff_r={\bf0}_m$, i.e. repulsive forces are no longer required
 \item $\Delta\bfsigma_{r}(t)={\dot\Delta\bfsigma}_{r}(t)={\bf0}_m$, i.e. the transient vanished
\end{enumerate}
the cooperative task can be restored (see Figure~\ref{fig:planner_scheme}) starting from the condition $\bfsigma_r=\bfsigma_r(t_s)$. 
\section{Simulation case study}\label{sec:case_study}
In this simulation case study, a setup composed of $N=3$ mobile arms is considered and is depicted in Figure~\ref{fig:setup}. In detail, each worker robot is a Comau Smart Six ($6$-DOFs) mounted on holonomic mobile base in order to move in the $xy$ plane ($2$-DOFs); the team's goal is to cooperatively transport loads from the picking conveyor belts on the left of the cell to the deposit one on the right. Such a task can be easily formulated by means of absolute and relative task variables introduced in~\eqref{eq:task_tot} where the former are appropriate for expressing the position and orientation of the grasped object,  while the latter  for expressing robots formation. Simulation results are provided in the video available at the following link\footnote{www.automatica.unisa.it/video/CoopHumanSafetySYROCO.mp4}.
\begin{psfrags}
\def\scal{0.8}   
\def\scalsmall{0.6}
 \psfrag{r1}[cc][][\scalsmall]{ $\text{Wo}1$ }
 \psfrag{r2}[cc][][\scalsmall]{ $\text{Wo}2$ }
 \psfrag{r3}[cc][][\scalsmall]{ $\text{Wo}3$ }
 \psfrag{bs1}[cc][][\scalsmall]{ $\text{BS}1$ }
 \psfrag{Cam1}[cc][][\scalsmall]{\hspace{0.1cm}}
 \psfrag{Cam2}[cc][][\scalsmall]{\hspace{0.1cm}}
 \psfrag{bs2}[cc][][\scalsmall]{ $\text{BS}2$ }
 \psfrag{pick1}[cc][][\scalsmall]{ $\text{PS}1$ }
 \psfrag{pick2}[cc][][\scalsmall]{ $\text{PS}2$ }
 \psfrag{dep}[cc][][\scalsmall]{ $\text{DS}$ }
 \psfrag{sigw}[cc][][\scalsmall]{ $\Sigma_w$ }
\mypsfrag{7.5}{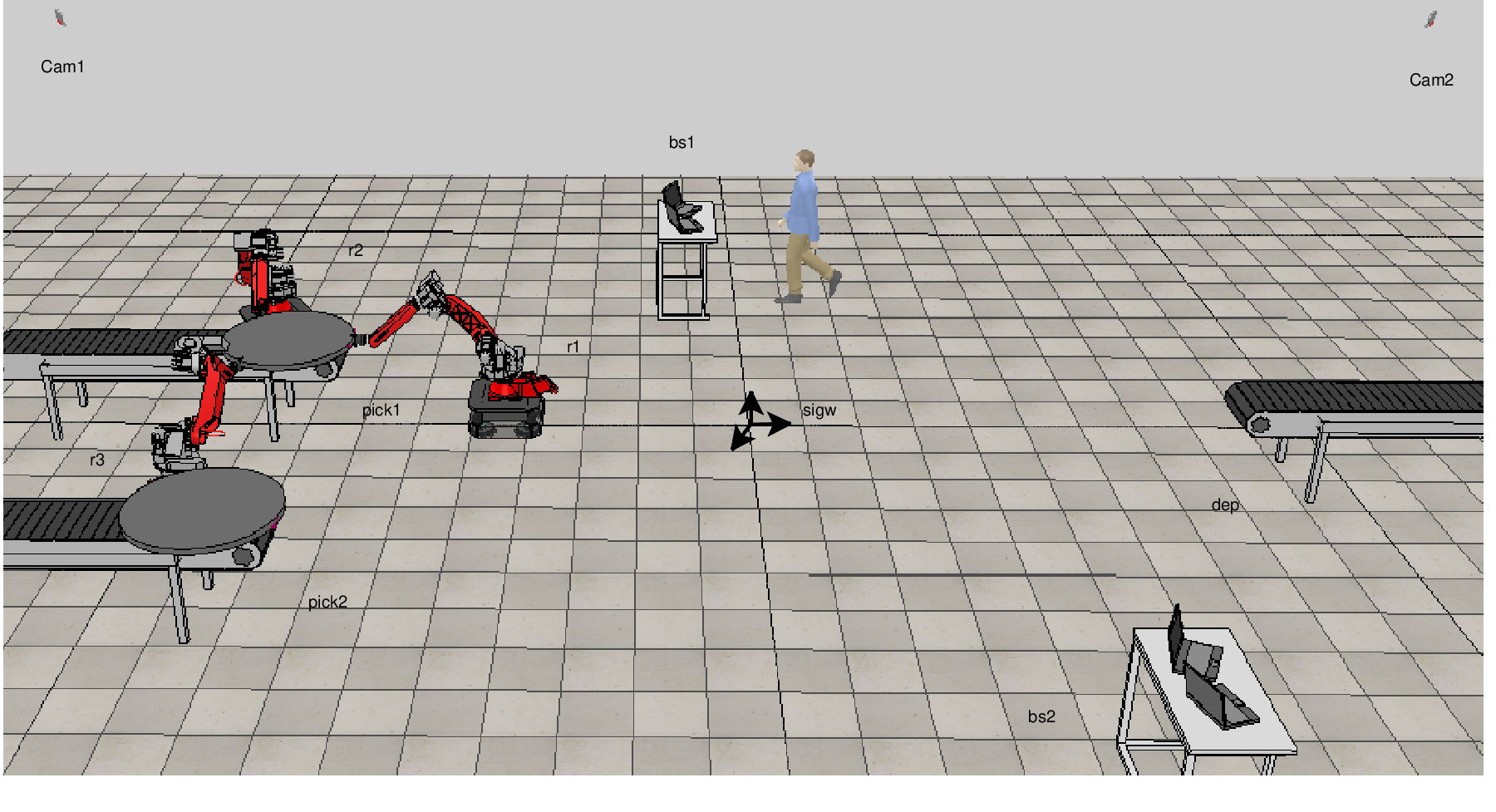}{-12pt}{Cell configuration composed by $3$ workers ($\text{Wo}i,\, i=1,2,3$), load picking and depositing stations ($\text{PS}i,\,i=1,2$ and $\text{DS}$, respectively) and base stations for  operators ($\text{BS}i,\,i=1,2$); $\Sigma_w$ is the world reference frame. }{fig:setup} 
\end{psfrags} 
\\Moreover, in order to asses the level of human safety, 
 the coefficients of $f$ in~\eqref{eq:f} are chosen in compliance with Properties $1$ and $2$ as
\begin{equation}\label{eq:f_coeff_case}
\left\{
\begin{aligned}
\alpha_1(d)&=k_1 d \\
\alpha_2(\dot d)&=k_2\tanh(\dot d)
\end{aligned}
\right.
\end{equation}
with $k_1,k_2\in\Re^+$, leading to
\begin{equation}\label{eq:f_case}
f(\bfp,\dot \bfp,\bfp_o,\dot \bfp_o ) = k_1 d+k_2\tanh(\dot d)
\end{equation}
where $\bfp_o$ is selected as the chest position of the human operator. 
The ratio behind the expression of function $f$ is that it is a combination of a linear term with respect to the distance $d$ and a  monotonically increasing term with respect to the derivative $\dot d$ whose contribution to the safety $f$ is negative when distance is decreasing ($\dot d<0$) and positive  for increasing values ($\dot d>0$). Starting from the expression of $f$ in~\eqref{eq:f_case}, $F^i$ ($i=1,2,3$) and $F$ in~\eqref{eq:fc} are computed according to the procedure in Section~\ref{sec:humansafetyassess}. The computation of the value $F_{min}$ such that to ensure $d\geq d_{min}$ in Remark~\ref{rm:fmin} is shown in the Appendix for the selected safety function in~\eqref{eq:f_case}. 
Moreover, due to the redundancy of the robots at hand, i.e. $n_i=8$ and $p=6$, the vector of joint accelerations $\ddot \bfq_{n,i}$ in~\eqref{eq:jointinput} can be locally exploited to maximize the $i$\ts th safety index $F^i$. To this aim, the acceleration vector is designed as in~\cite{Hsu_ACC1988}, which is standard for second order kinematics, while the desired velocity to be projected in the null space of $\bfJ_i$ is computed with a gradient technique accordingly to the procedure in~\cite{Zanchettin_TRO2013}.
\\ The following gains are selected for the virtual input in~\eqref{eq:jointinput}, $k_{\sigma}=20$, $\lambda_{\sigma}=100$,  while the following ones for the avoidance strategy, $k_d=4.5$, $k_p=5$, $\bfM=\bfI_{18}$, $ \bfD=6.5\bfI_{18}$, $\bfK=10\bfI_{18}$ $k_{r}=15$, $\Delta F=15$ in equations~\eqref{eq:scaling_dyn}, \eqref{eq:delta_impedance} and \eqref{eq:rep_force}, respectively; finally, the minimum cumulative safety value is set as $F_{min}=80$.
\\Figure~\ref{fig:snap} shows key snapshots of the simulation while detailed results are presented in Figures~\ref{fig:safetyF}-\ref{fig:nr_pos}.
In the simulation study, human behaviour mainly interferes in two phases with the robots task leading once to modify the nominal trajectory via velocity scaling and once via path deformation. In detail, the scaling phase (from $t=8.7$\ts s up to $t=20$\ts s) occurs when the robots move towards the operator standing at the first base station (Figure~\ref{fig:snap}.b), while the impedance phase 
(from  $t=39.2$\ts s up to $t=45.1$\ts s) occurs when the operator crosses the robots nominal path and the scaling trajectory is no longer sufficient for ensuring minimum safety (Figure~\ref{fig:snap}.c). 
\vspace{0.1cm}
\begin{psfrags}
\def\scal{0.8}  
 \psfrag{a}[cc][][\scal]{ a)}
 \psfrag{b}[cc][][\scal]{ b)}
 \psfrag{c}[cc][][\scal]{ c)}
\mypsfrag{8.3}{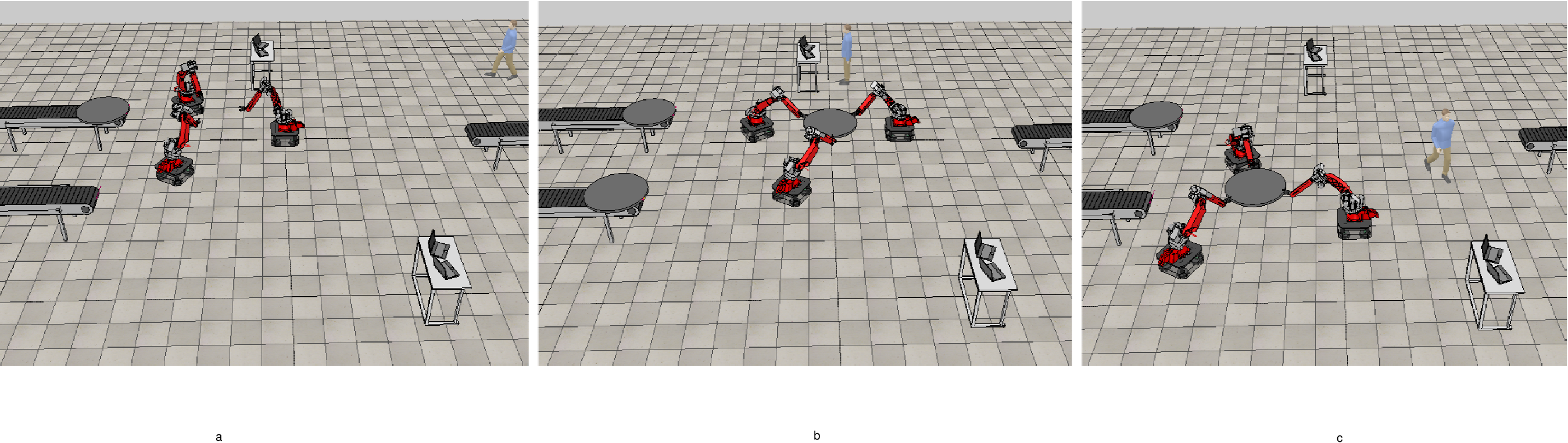}{-5pt}{Simulation snapshots representing the initial system configuration (a), the scaling phase (b) and the impedance phase (c), respectively.}{fig:snap} 
\end{psfrags}
\\Figure~\ref{fig:safetyF} shows the progress of the safety index during the human-robot interaction; in particular, it makes evident that, during the scaling phase ($S$), the safety value is almost saturated at its minimum value while an increase of it is detected at the beginning  of the impedance phase ($I$) due to the path constraint relaxation. 
\begin{psfrags}
\def\scal{0.8}  
\def\scallegend{0.6}  
\def\scalnum{0.6}
 \psfrag{t}[cc][][\scal]{ $time$ [s]}
 \psfrag{y}[cc][][\scal]{}
 \psfrag{f}[cc][][\scallegend]{ $F$}
 \psfrag{fmin}[cc][][\scallegend]{ $F_{min}$}
 \psfrag{S}[cc][][\scal]{$S$}
 \psfrag{I}[cc][][\scal]{$I$}
 \psfrag{0}[cc][][\scalnum]{0}
 \psfrag{10}[cc][][\scalnum]{10}
 \psfrag{20}[cc][][\scalnum]{20}
 \psfrag{30}[cc][][\scalnum]{30}
 \psfrag{40}[cc][][\scalnum]{40}
 \psfrag{50}[cc][][\scalnum]{50}
 \psfrag{60}[cc][][\scalnum]{60}
 \psfrag{70}[cc][][\scalnum]{70}
 \psfrag{100}[cc][][\scalnum]{100}
 \psfrag{200}[cc][][\scalnum]{200}
 \psfrag{300}[cc][][\scalnum]{300}
 \psfrag{4}[cc][][\scalnum]{4}
 \psfrag{8}[cc][][\scalnum]{8}
 \psfrag{12}[cc][][\scalnum]{12}
 \psfrag{16}[cc][][\scalnum]{16}
\mypsfrag{8.5}{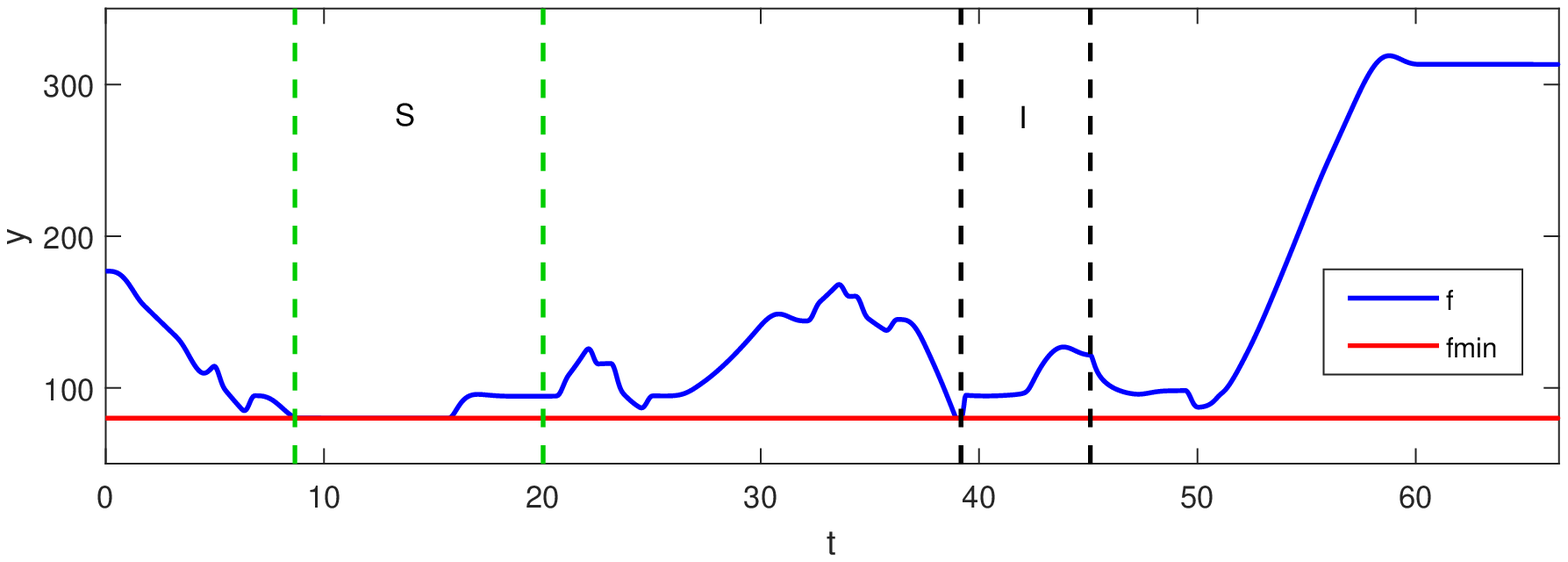}{-8pt}{Evolution of the cumulative safety function (in blue) with respect to its minimum allowed value (in red); scaling and impedance phases are marked with $S$ and $I$, respectively. }{fig:safetyF} 
\end{psfrags}

Concerning the scaling strategy, Figure~\ref{fig:deltas} shows how the scaling parameters vary over time; in particular, scaling phase is firstly characterized by a decrease of $\dot \Delta s$, i.e. a slowdown of nominal trajectory, and then an increase of it in order to restore nominal trajectory tracking ( i.e. the condition $\ddot \Delta s=\dot \Delta s=\Delta s=0$). This effect is also evident from Figure~\ref{fig:nr_pos} where the centroid position of the nominal trajectory is compared to that of the reference one.  
Moreover, Figure~\ref{fig:nr_pos} shows how trajectory is modified when path constraint is relaxed; in this case, starting from ${\bfsigma_r(t_s=39.2\,\text{s})}$ the reference trajectory evolves according to the impedance model in~\eqref{eq:delta_impedance} and then, when repulsive action is no longer necessary, it returns again to ${\bfsigma_r(t_s)}$ in order to restore the tracking of the nominal trajectory. 

\begin{psfrags}
\def\scal{0.8}   
\def\scalnum{0.6}
 \psfrag{t}[cc][][\scal]{ $time$ [s]}
 \psfrag{imp}[cc][][\scal]{\hspace{0.2cm}$I$}
 \psfrag{y1}[cc][][\scal]{ $\Delta_s$}
 \psfrag{y2}[cc][][\scal]{ $\dot \Delta_s$}
 \psfrag{y3}[cc][][\scal]{ $\ddot \Delta_s$}
 \psfrag{S}[cc][][\scal]{ \hspace{-0.1cm}$S$}
 \psfrag{I}[cc][][\scal]{}
 \psfrag{0}[cc][][\scalnum]{0}
 \psfrag{10}[cc][][\scalnum]{10}
 \psfrag{20}[cc][][\scalnum]{20}
 \psfrag{-20}[cc][][\scalnum]{-20}
 \psfrag{30}[cc][][\scalnum]{30}
 \psfrag{40}[cc][][\scalnum]{40}
 \psfrag{50}[cc][][\scalnum]{50}
 \psfrag{60}[cc][][\scalnum]{60}
 \psfrag{70}[cc][][\scalnum]{70}
 \psfrag{-0.5}[cc][][\scalnum]{-0.5}
 \psfrag{-1}[cc][][\scalnum]{-1}
 \psfrag{-2}[cc][][\scalnum]{-2}
 \psfrag{2}[cc][][\scalnum]{2}
 \psfrag{-4}[cc][][\scalnum]{-4}
 \psfrag{-10}[cc][][\scalnum]{-10}
\mypsfrag{8.}{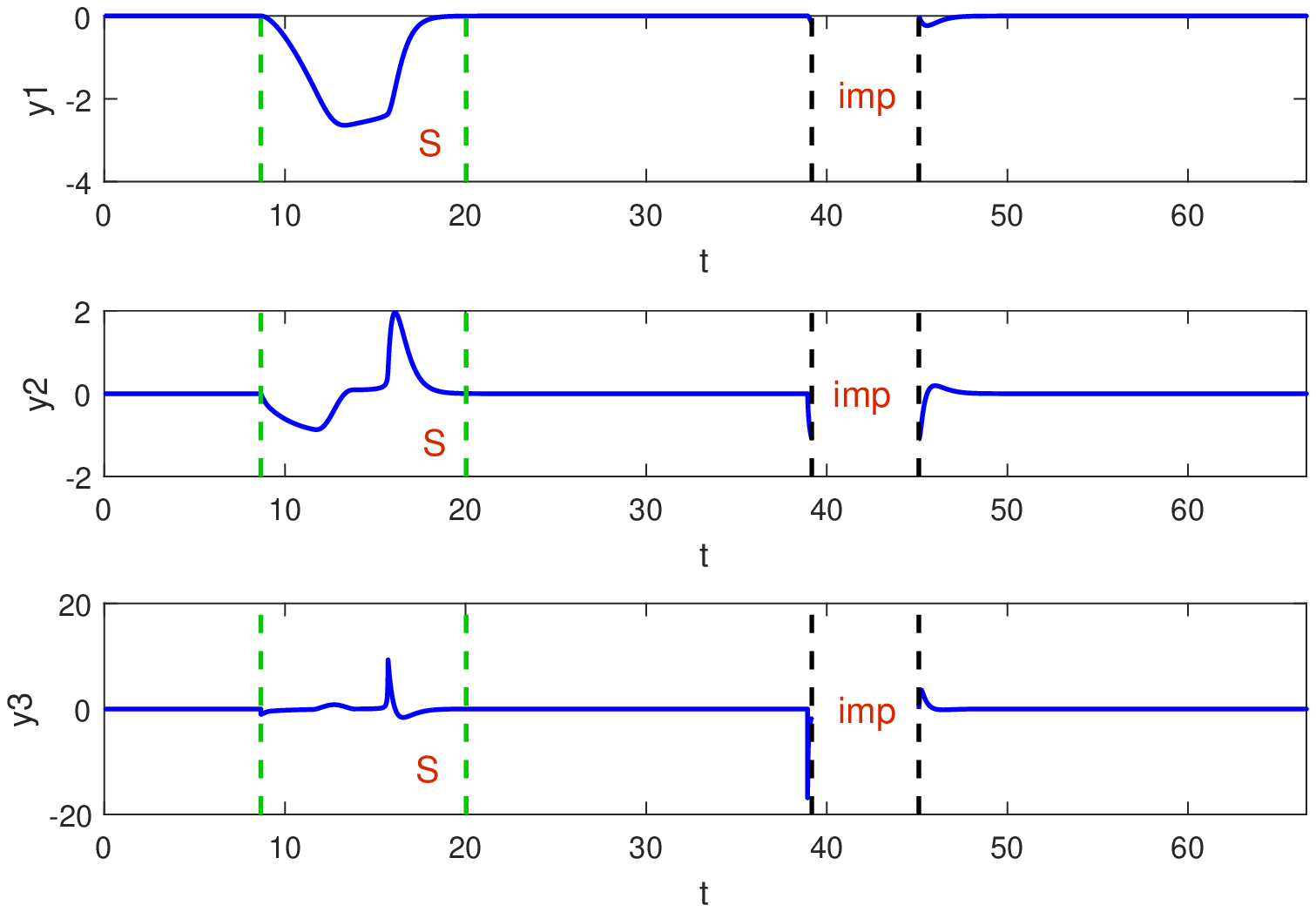}{-9pt}{Evolution of scaling terms; scaling and impedance phases are marked with $S$ and $I$, respectively. In the impedance phase, no plots are provided since the nominal path is abandoned.}{fig:deltas} 
\end{psfrags}

\begin{psfrags}
\def\scal{0.8}  
\def\scallegend{0.6}  
\def\scalnum{0.6}
 \psfrag{t}[cc][][\scal]{ $time$ [s]}
 \psfrag{y1}[cc][][\scal]{ $\bfsigma_{1,x}$}
 \psfrag{y2}[cc][][\scal]{ $\bfsigma_{1,y}$}
 \psfrag{y3}[cc][][\scal]{ $\bfsigma_{1,z}$}
 \psfrag{S}[cc][][\scal]{$S$}
 \psfrag{I}[cc][][\scal]{$I$}
 \psfrag{pn1}[cc][][\scallegend]{ $n$}
 \psfrag{pr1}[cc][][\scallegend]{ $r$}
 \psfrag{pn2}[cc][][\scallegend]{ $n$}
 \psfrag{pr2}[cc][][\scallegend]{ $r$}
 \psfrag{pn3}[cc][][\scallegend]{ $n$}
 \psfrag{pr3}[cc][][\scallegend]{ $r$}
 \psfrag{0}[cc][][\scalnum]{0}
 \psfrag{10}[cc][][\scalnum]{10}
 \psfrag{20}[cc][][\scalnum]{20}
 \psfrag{-20}[cc][][\scalnum]{-20}
 \psfrag{30}[cc][][\scalnum]{30}
 \psfrag{40}[cc][][\scalnum]{40}
 \psfrag{50}[cc][][\scalnum]{50}
 \psfrag{60}[cc][][\scalnum]{60}
 \psfrag{70}[cc][][\scalnum]{70}
 \psfrag{-5}[cc][][\scalnum]{-5}
 \psfrag{5}[cc][][\scalnum]{5}
 \psfrag{1}[cc][][\scalnum]{1}
 \psfrag{0.8}[cc][][\scalnum]{0.8}
 \psfrag{0.6}[cc][][\scalnum]{0.6}
\mypsfrag{8.}{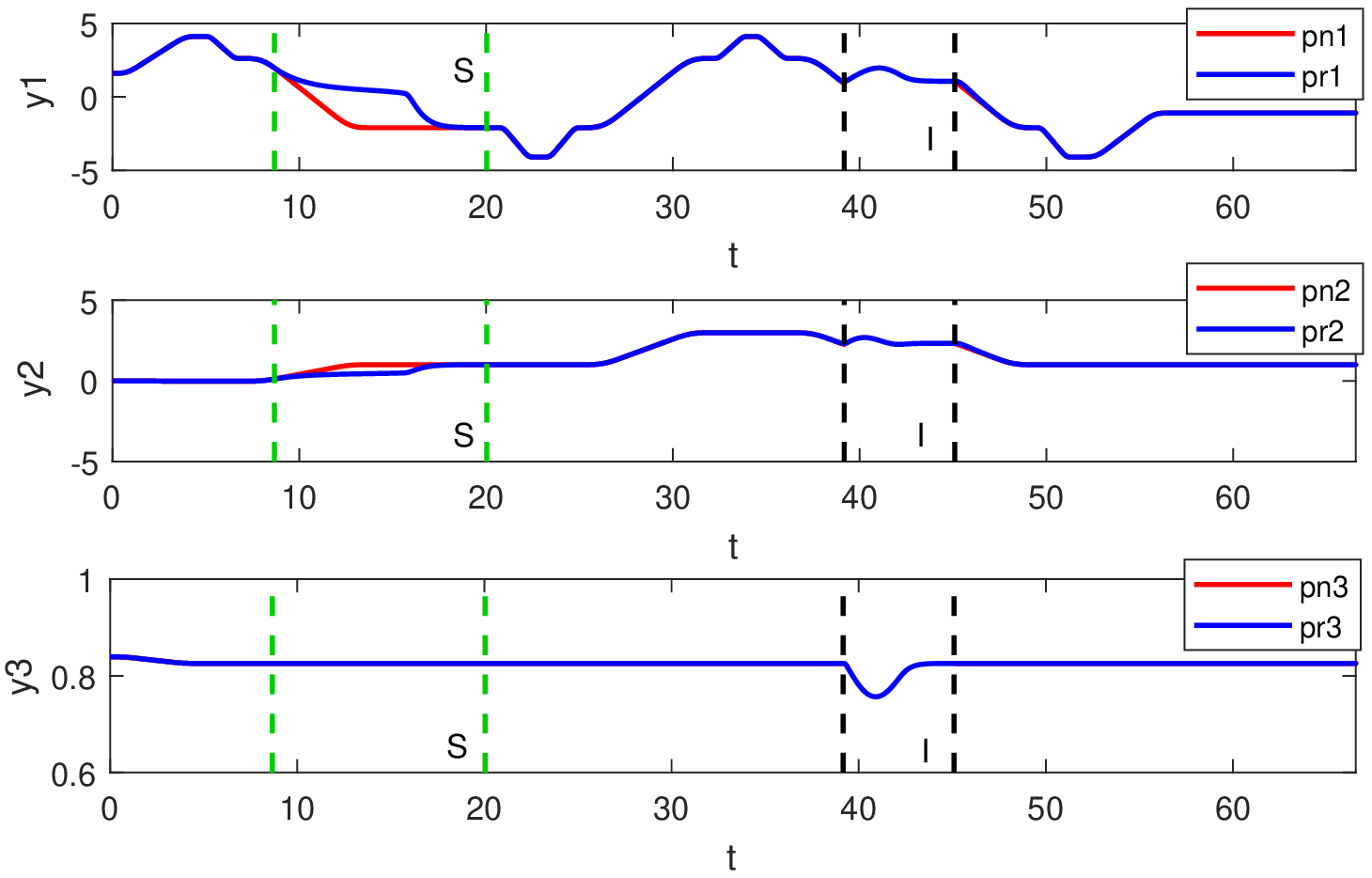}{-9pt}{Evolution of the nominal ($n$, in red) and reference ($r$, in blue) trajectories of the team centroid position; scaling and impedance phases are marked with $S$ and $I$, respectively. In the impedance phase, nominal trajectory is not shown since its tracking is aborted.}{fig:nr_pos} 
\end{psfrags}
%
%

\section{Conclusions}\label{sec:conclusions}
In this work, a general approach to achieve cooperative tasks by multi-robot systems in coexistence with human operators was presented.
To this aim, the human-robot safe interaction is first assessed by the definition of a general safety index and, then, a strategy capable of ensuring  a safe human-robot interaction is defined.
At the same time, this strategy is such as  to preserve as much as possible the desired cooperative task and, only in the case the human safety cannot be longer ensured, the task is aborted and a suitable avoidance strategy is undertaken.
As future work, the approach will be extended to cope with a decentralized architecture and will be validated through experiments on a real setup.

\appendix
\setcounter{equation}{0}
\renewcommand\theequation{A.\arabic{equation}}
\section*{Appendix}\label{app:min_fc}
For a given value $d_{min}$ of $d$ in \eqref{eq:dmin}, the objective is to compute $F_{min}$ such as
$F\ge F_{min}$ implies $d\ge d_{min}$.
For this purpose, let us first consider the case of a single $n_l$ link manipulator. 
The required $F_{min}$ can be computed as the maximum value of $F$ for all $\bfp_o$ such as  $d=d_{min}$
\begin{equation}\label{eq:bound_single_robot}
\begin{aligned}
F
&=\sum_{l=1}^{n_l}\int_0^1 \left(k_1d_{l,r}+k_2\tanh(\dot d_{l,r})\right)~dr \\
&\leq \sum_{l=1}^{n_l} \left( k_1\int_0^1 \|\bfp_{l,0}+r(\bfp_{l,1}-\bfp_{l,0})-\bfp_o\|~dr +k_2 \right)\\
&\leq k_1\left(\frac{1}{2}\sum_{l=1}^{n_l} L_l+ \sum_{l=1}^{n_l} \|\bfp_{l,0}-\bfp_o\|\right)+k_2\,n_l  \\
\end{aligned}
\end{equation}
where $L_l$ is the length of the $l$\ts th link.
Let $\bfp^\star$ be the generic point on the robot structure at distance $d_{min}$ from $\bfp_o$, i.e., $\|\bfp_o-\bfp^\star\|=d_{min}$, 
then the following inequalities  holds\\
\begin{equation}\label{eq:dist_single_robot}
\begin{aligned}
\|\bfp_{l,0}-\bfp_o\pm \bfp^\star\|
&\leq \|\bfp_{l,0}-\bfp^\star\|+\|\bfp_o-\bfp^\star\| \\
&\leq \sum_{l=1}^{n_l} L_l + d_{min} = L+d_{min}
\end{aligned}
\end{equation}
where $L=\sum_{l=1}^{n_l} L_l$ and the obvious relation $\|\bfp_{l,0}-\bfp^\star\|\leq L$ has been exploited.
Hence, in virtue of \eqref{eq:bound_single_robot} and \eqref{eq:dist_single_robot}, it follows that for single robot 
\begin{equation*}
F_{min}=k_1\left(\frac{2n_l+1}{2}L + n_l\, d_{min}\right)+k_2\,n_l \\
\end{equation*}
By generalizing to the case of $N$ robots, it holds 
\begin{equation}\label{eq:bound_team_robot}
\begin{aligned}
F&\leq\sum_{i=1}^{N} \left[ k_1\left(\frac{1}{2}L^i+ \sum_{l=1}^{n_{l}^i+1} \|\bfp^i_{l,0}-\bfp_o\|\right)+k_2\,(n_{l}^i+1) \right]\\
\end{aligned}
\end{equation}
where $L^i=\sum_{l=1}^{n_l^i+1} L_l^i$ also takes into account the virtual link to the team centroid.
Moreover, since the robots are assumed to be in formation, the structure composed by the $i$\ts th manipulator and the one at minimum distance $d_{min}$ can be analysed in turn as an ``aggregate" manipulator whose maximum sum of link lengths is $L^i + L_{max}$ with $L_{max}=\max_{i\in{1,..,N}}L^i$; therefore, it holds 
\begin{equation}\label{eq:dist_team_robot}
\|\bfp_{l,0}^i-\bfp_o\| \leq L^i + L_{max} + d_{min}
\end{equation}
From \eqref{eq:bound_team_robot} and \eqref{eq:dist_team_robot}, it follows that 
\begin{equation*}
\begin{aligned}
F_{min}\!=\!\!\sum_{i=1}^{N}&\,[k_1\!\left(\frac{2n_{l}^i+3}{2}L^i\!+\! (n_{l}^i +1)(L_{max}+ d_{min})\right)\\ &\!\!+\!k_2 \,(n_{l}^i+1)  ]
\end{aligned}
\end{equation*}
is such to ensure that each point on each manipulator is at least at distance $d_{min}$ from the operator.

\bibliography{biblio}
\end{document}